\documentclass[lettersize,journal]{IEEEtran}
\usepackage{amsmath,amsfonts}
\usepackage{algorithmic}
\usepackage{algorithm}
\usepackage{array}
\usepackage[caption=false,font=normalsize,labelfont=sf,textfont=sf]{subfig}
\usepackage{textcomp}
\usepackage{stfloats}
\usepackage{url}
\usepackage{verbatim}
\usepackage{graphicx}
\usepackage{cite}
\usepackage{times}
\usepackage{epsfig}
\usepackage{graphicx}
\usepackage{amssymb}
\usepackage{fontawesome}
\begin{document}

\title{Mixed Hierarchy Network for Image Restoration}

\author{Hu Gao, Depeng Dang
\thanks{Hu Gao and Depeng Dang are with the School of
Artificial Intelligence, Beijing Normal University,
Beijing 100000, China (e-mail: gao\_h@mail.bnu.edu.cn, ddepeng@bnu.edu.cn).}}


\IEEEpubid{}

\maketitle

\begin{abstract}
Image restoration is a long-standing low-level vision problem, e.g., deblurring and deraining. In the process of image restoration, it is necessary to consider not only the spatial details and contextual information of restoration to ensure the quality but also the system complexity. Although many methods have been able to guarantee the quality of image restoration, the system complexity of the state-of-the-art (SOTA) methods is increasing as well. Motivated by this, we present a mixed hierarchy network that can balance these competing goals. Our main proposal is a mixed hierarchy architecture, that progressively recovers contextual information and spatial details from degraded images while we use simple blocks to reduce system complexity. 
Specifically, our model first learns the contextual information at the lower hierarchy using encoder-decoder architectures, and then at the higher hierarchy operates on full-resolution to retain spatial detail information. 
Incorporating information exchange between different hierarchies is a crucial aspect of our mixed hierarchy architecture. To achieve this, we design an adaptive feature fusion mechanism that selectively aggregates spatially-precise details and rich contextual information.
In addition, we propose a  selective multi-head attention mechanism  with linear time complexity as the middle block of the encoder-decoder to adaptively retain the most crucial attention scores. 
What's more, we use the nonlinear activation free block as our base block to reduce the system complexity.
The resulting tightly interlinked hierarchy architecture, named as MHNet, delivers strong performance gains on several image restoration tasks, including image deraining, and deblurring. 
\end{abstract}


\section{Introduction}
\IEEEPARstart{I}{mage} restoration, such as image deblurring and image deraining is a family of inverse problems for obtaining a high-quality image from a corrupted input image. The image restoration is generally modeled as follows: 
\begin{equation}
	\label{equ:01}
	l(x,y)=h(x,y) \cdot g(x,y)+ \sigma(x,y)
\end{equation}

Where $l(x,y), h(x,y)$ denote an observed low-quality image, and its corresponding high-quality image, respectively. And $g(x,y), \sigma(x,y)$ denote the degradation function and the noise during the imaging and transmission processes, respectively.  Image restoration aims to enhance the visual quality of images $l(x,y)$ and get the high-quality image $h(x,y)$. It is typically an ill-posed inverse problem since there are many candidates for any original input and the image degradation procedures are irreversible. To restrict the infinite feasible solutions space to natural images, traditional methods~\cite{1992Nonlinear,  2002Scale, 2005Fields, 2011Image, 2011Single} explicitly design domain-relevant priors, task-relevant priors, and various constraints for the given kind of restoration problem. Then, with the appropriately designed priors, the potential high-quality image $\hat{x}$ can be obtained by solving a MAP(maximum a posteriori) problem:
\begin{equation}
	\label{equ:02}
	\hat{x}= \underset {x} { \operatorname {arg\,max}} \log p(y|x) + \log p(x).
\end{equation}

However, designing such priors is a challenging task and often not generalizable.  Recently, stimulated by the success of deep learning for high-level vision tasks, numerous deep models~\cite{dai2019second,2020Residual,Zamir2021MPRNet,Zamir2022MIRNetv2,chen2022simple,chu2022nafssr,Wang_2022_CVPR,Chen_2021_CVPR} have been developed to tackle low-level vision tasks. With large-scale data, deep models such as Convolutional Neural Networks(CNNs) and Transformer can implicitly learn more general priors by capturing natural image statistics and achieving state-of-the-art(SOTA) performance in image restoration.
\begin{figure}[t] 
	\centering
	\includegraphics[width=0.5\textwidth]{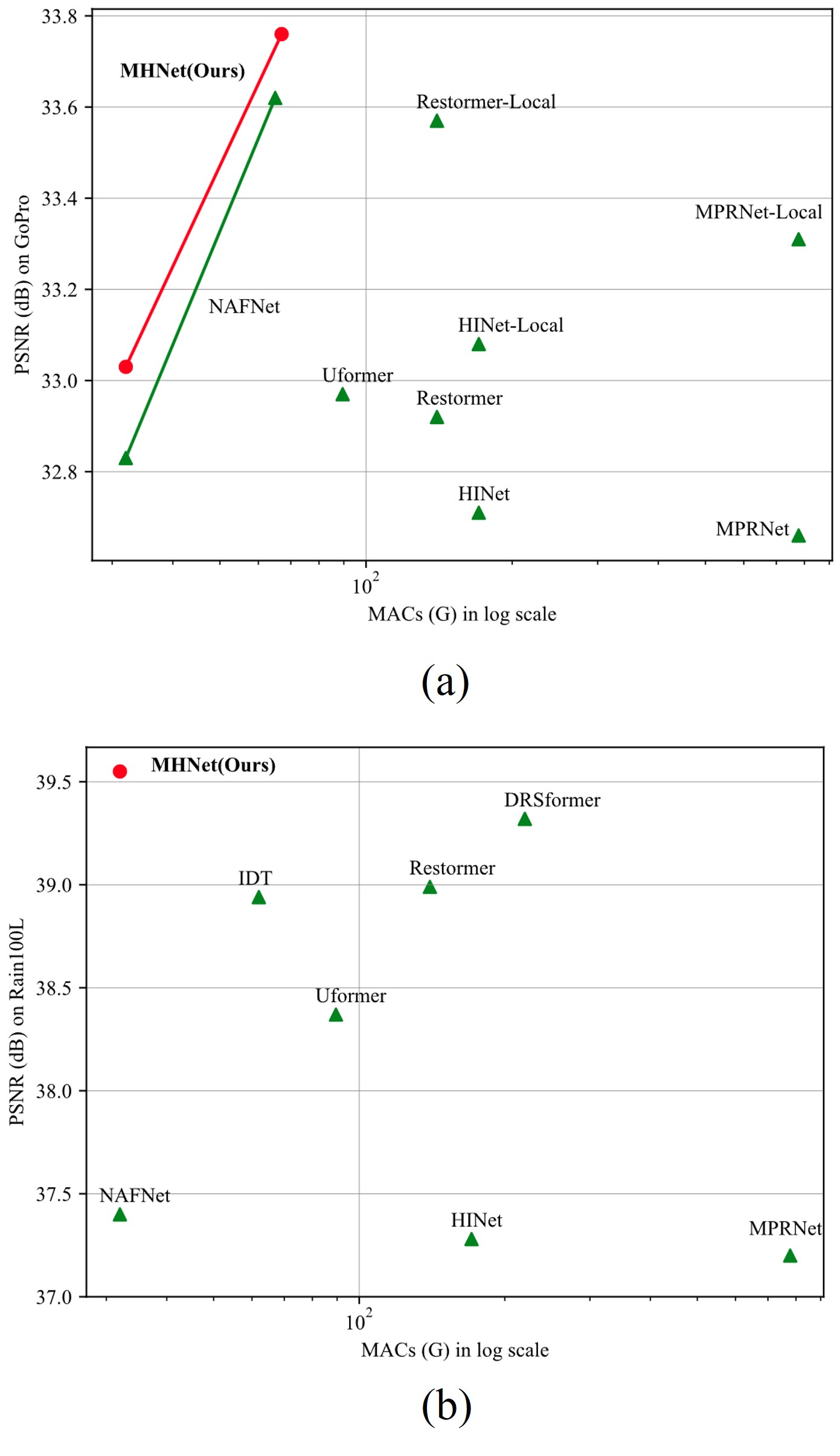}
	\caption{Computational cost vs. PSNR of models on Image Deblurring (top) and Image Deraining (bottom) tasks. Our MHNet achieve the SOFT performance with up to 85\% of cost reduction on Image Deraining .}
	\label{fig:param}
\end{figure}

The performance gain of deep learning methods over the others is primarily attributed to its model design. Numerous network modules and functional units for image restoration have been investigated to design various models in recent years, including encoder-decoder~\cite{chen2022simple,chu2022nafssr}, multi-stage network~\cite{Zamir2021MPRNet, Chen_2021_CVPR,PREnet,RESCAN}, dual network~\cite{2018LearningD, 2022Learning,2020Refining,chen2020decomposition}, recursive residual~\cite{zhang2018image,2019Real} and various transformer improvements~\cite{zhang2023accurate, Zamir2021Restormer, Tsai2022Stripformer,Wang_2022_CVPR}. Despite their good performance, these methods suffer from high system complexity, it often takes a long time and a lot of video memory to run a model with a large number of parameters. 

Based on the above facts, a natural question arises: Is it possible to summarize the advantages of the above network structure and propose a neural network with computational costs to get considerable results? To this end, we analyze the characteristics of the above network structure and identify the architectural bottlenecks that hamper their performance. First, the structure of the encoder-decoder can learn the contextualized features but is unreliable in preserving spatial image details. Second, the multi-stage network can learn multi-scale contextual information and preserve fine spatial details, while may yield sub-optimal results. Third, the dual network is more flexible for low-level vision tasks, yet the results are not good. What's more, in order to avoid gradient disappearance, many network structures use the  recursive residual but only as a sub-network. Finally, despite showing outstanding performance, existing Transformer backbones for image restoration still suffer from serious defects, such as self-attention computation. 

To address the above issues, we propose a mixed hierarchy image restoration network architecture, named as MHNet. Our approach is universal, as it can be adapted for generic image restoration problems. Specifically, (1) We draw lessons from the idea of multi-stage and dual networks and use multiple subnetworks to extract features, the first hierarchy uses an encoder-decoder to learn multi-scale contextual information, while the last hierarchy maintains fine spatial details by operating on the full-resolution. 
(2) In order to incorporate information exchange between different hierarchies, we design an adaptive feature fusion mechanism (AFFM) to selectively aggregates spatially-precise details and complementary contextual information. 
(3) We learn the attention mechanism of the transformer and propose a selective multi-head attention mechanism (SMAM) to adaptively retain the most crucial attention scores, thereby facilitating better feature aggregation. As shown in Figure~\ref{fig:param}, our MHNet model achieves state-of-the-art performance while being computationally efficient in comparison to existing methods.

The main contributions of this work are:
\begin{enumerate}
	\item By analyzing the SOTA methods and extracting their essential components, we proposed a novel mixed hierarchy method that efficiently generates a restored image that is rich in contextual and accurate in spatial detail.
    \item We design an adaptive feature fusion mechanism (AFFM) across hierarchy that effectively combines spatially-precise details and complementary contextual information.
	\item We propose a selective multi-head attention mechanism (SMAM) as the encoder-decoder middle block that is capable of aggregating the  most crucial information contained in the feature maps captured by convolution.
    \item We use the nonlinear activation free block as our base block to reduce the system complexity.
    \item We demonstrate the versatility of MHNet by setting new state-of-the-art on 6 synthetic and real-world datasets for various restoration tasks (image deraining and deblurring) while maintaining low complexity. Further, we provide detailed analysis, qualitative results, and generalization tests.
\end{enumerate}
\section{Related Work}
\label{sec:1}

\subsection{Image Restoration}
Image restoration aims to restore a degraded image to a clean one, it is split into a large number of sub-problems, for image deblurring, and image deraining among others. Early restoration approaches are  usually defined by hand-crafted priors that narrowed the ill-posed nature of the problems by reducing the set of plausible solutions, such as total variation~\cite{1992Nonlinear,Chan1998TotalVB}, sparse coding~\cite{2015Removing,20060K,2007Sparse} self-similarity~\cite{2005A,2007Image}, gradient prior~\cite{2008High,2013Unnatural}, etc. With the rapid development of deep learning, numerous works based on deep learning have gained great popularity for image restoration, and achieve SOTA results~\cite{Zamir2021MPRNet,zhang2023accurate,Zamir2021Restormer,Chen_2021_CVPR,Tsai2022Stripformer,chu2021tlc}. Although the above methods achieve a considerable result, they suffer from high system complexity and often takes a lot of video memory to run a model with a large number of parameters.

\subsection{Encoder-Decoder Approaches} 
In recent years, encoder-decoder have gained great attention from researchers in the field of image restoration. There are many variations of encoder-decoder, such as using multiple layers of convolution and deconvolution operators~\cite{2016Imagenip}, cascading of the network input into intermediate layers~\cite{Mastan2019MultiLevelEA}, many compositions of the encoder-decoder sub-networks~\cite{2022As}, deepen the  network depth, and introducing the capsule network~\cite{2020RedCap}. By learning end-to-end mappings from corrupted images to the original ones, this methods~\cite{Zamir2021MPRNet,Zamir2021Restormer,2020arXiv201215028C} achieve SOTA results on image restoration. However, the encoder-decoder approach is not reliable in preserving spatial image details.

\subsection{Multi-Stage Approaches} 
Multi-stage approaches divide the task of image restoration into multiple stages by employing light-weight subnetworks~\cite{2018Scale,2018Lightweight,RESCAN,PREnet,Zamir2021MPRNet,zhang2022event,Zhang_2019_CVPR}. In this way, spatial details and high-level contextualized information is gradually captured, the image restoration process becomes more controlled. However, the multi-stage model parameters are too large, the MACs are low, may yield sub-optimal results, and the loss of $nan$ is easy to occur. 

\subsection{Dual Networks} 
 Dual Networks have two parallel branches, which respectively estimate the structure and detail components of the target signals from the input, and then reconstruct the final results according to the specific tasks formulation module. \cite{2018LearningD}first proposed the DualCNN framework, which has inspired the following work, including image dehazing~\cite{2018DehazeGAN,2019Dense,2019Dual}, image deraining~\cite{2018Fast}, image denoising~\cite{tian2021designing}, and image super-resolution/deblurring~\cite{2020Refining}. 

\subsection{Vision Transformer}
Due to the content-dependent global receptive field, the transformer architecture~\cite{2017Attention} has recently gained much popularity in the high-level computer vision community, such as image classification, segmentation, object detection~\cite{Conde_2021_CVPR,dosovitskiy2020vit,liu2021swin,chen2021outperform}. After the impressive performance on high-level vision tasks, researchers have been started to try to use it in image restoration~\cite{conde2022swin2sr,liang2021swinir,Zamir2021Restormer,Tsai2022Stripformer,Wang_2022_CVPR, DRSformer, kong2023efficient, IDT}.~\cite{liang2021swinir} proposed a strong baseline model named SwinIR. \cite{Zamir2021Restormer} also proposed an efficient Transformer model using a U-net structure named Restormer and achieved state-of-the-art results on several image restoration tasks. However, the state-of-the-art results rely on a large number of parameters and heavy computation (over 115.5M parameters), and high MACs, which are bad for devices with limited resources. 

\section{Method}
\begin{figure}[htb] 
	\centering
	\includegraphics[width=0.5\textwidth]{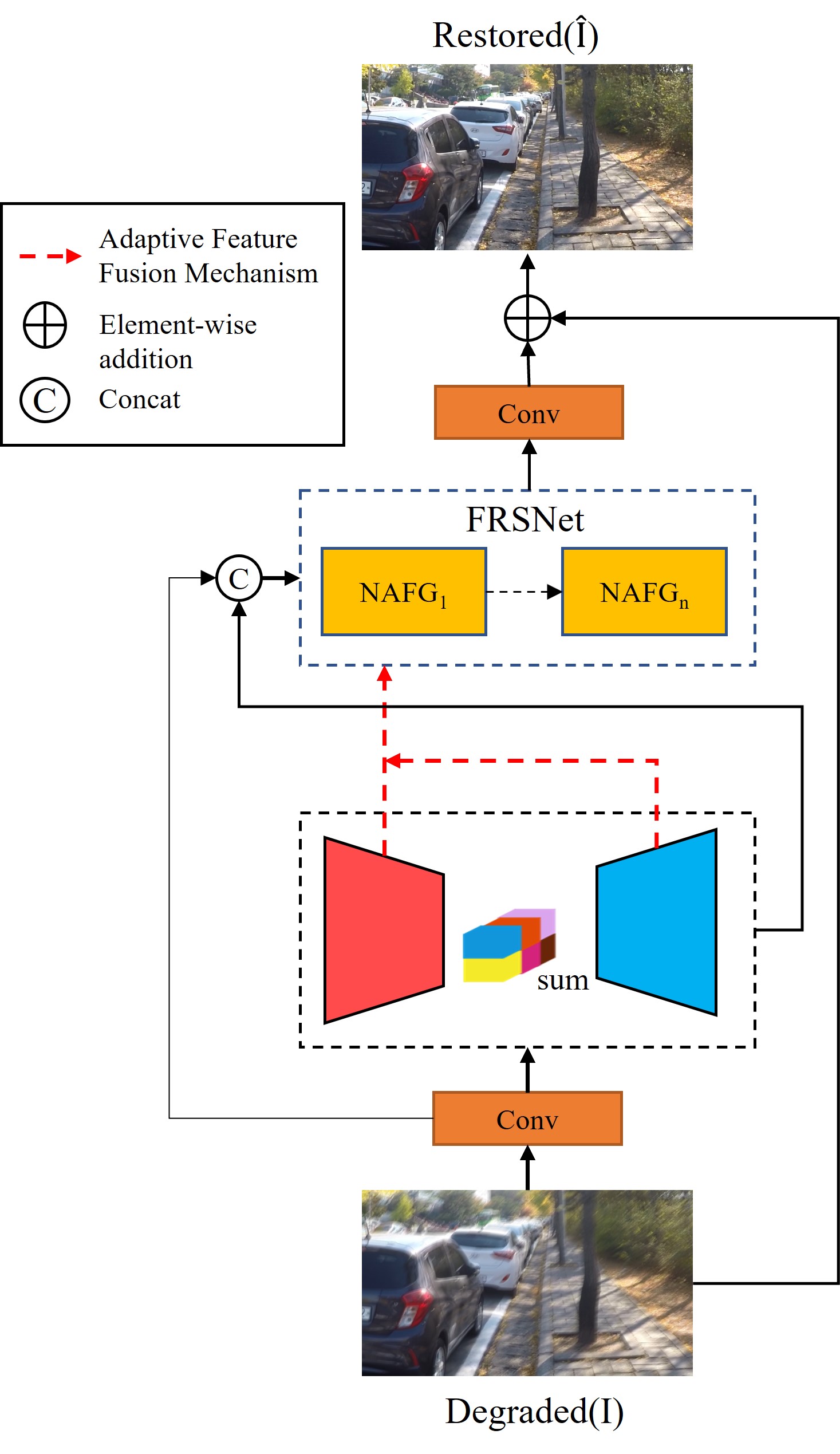}
	\caption{Architecture of MHNet for image restoration.
}
	\label{fig:network}
\end{figure}
Our original intention is to look for a network structure that summarizes the advantages of the existing network structure and gets considerable results with lower computational costs. The overview of the proposed framework MHNet for image restoration, shown in Figure.~\ref{fig:network}, consists of two hierarchy structures. The first hierarchy is based on encoder-decoder subnetworks that learn the contextualized features with broad context.  And the next hierarchy employs a subnetwork that operates on the full resolution to retain local information.

Instead of simply employing the U-Net as the encoder-decoder network, we're based on~\cite{chen2022simple} research, replacing or removing the nonlinear activation function by multiplication and using a simple network structure. What's more, we replace convolution with global self-attention in the middle block. Such a hybrid design can aggregate more information and let attention work on smaller resolutions generated by having convolutions do the spatial downsampling. Furthermore, we introduce a feature fusion mechanism to incorporate information exchange between different hierarchies.

\noindent\textbf{Overall Pipeline.} Given a degraded image $\mathbf{I} \in \mathbb R^{H \times W \times 3}$, MHNet first applies a convolution to obtain low-level features $\mathbf{X_{0}} \in \mathbb R^{H \times W \times C}$. Next, the $\mathbf{X_{0}}$ pass through a encoder-decoder network, yielding encoder features $\mathbf{E}[\mathbf{X_{e1}}, \mathbf{X_{e2}}, \mathbf{X_{e3}}, \mathbf{X_{e4}}]$ and decoder features $\mathbf{D}[\mathbf{X_{d1}}, \mathbf{X_{d2}}, \mathbf{X_{d3}}, \mathbf{X_{d4}}]$, respectively. In order to assist the image restoration, the encoder features $\mathbf{E}$ are added with the decoder features $\mathbf{D}$ via skip connections. Then, we input the features $\mathbf{X_1}$ of the concatenation of $\mathbf{X_0}$ and $\mathbf{X_{d4}}$ into $N$ number of full resolution  subnetwork (FRSNet), yielding deep features $\mathbf{X_d}\in \mathbb R^{H \times W \times C}$. It is worth noting that we introduce the feature fusion module between the encoder-decoder and FRSNet. Finally, we apply  convolution to the refined features to generate residual image $\mathbf{X}\in \mathbb R^{H \times W \times 3}$ to which degraded image is added to obtain the restored image:$\mathbf{\hat{I}} = \mathbf{X} +\mathbf{I}$. We optimize the proposed network using PSNR loss : 

\begin{equation}
	\label{equ:03}
	PSNR = 10 \cdot log_{10} \cdot \frac{(2^n-1)^2}{||\mathbf{\hat{I}}-\mathbf{\dot I}||^2}
\end{equation}

\noindent where $\mathbf{\dot I}$ denotes the ground-truth image.

\begin{figure*}[!htb] 
	\centering
	\includegraphics[width=1\textwidth]{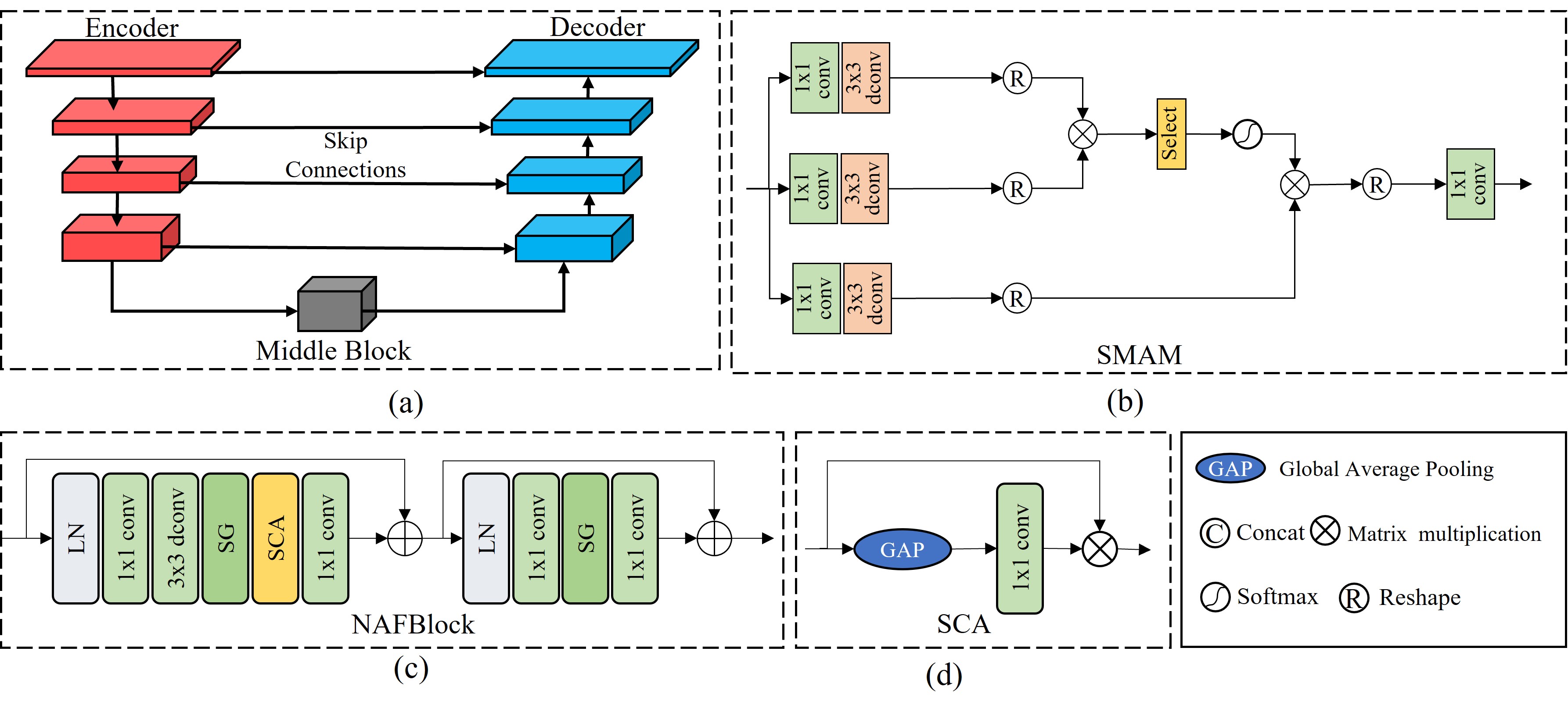}
	\caption{(a) Encoder-decoder subnetwork. (b) Selective multi-head attention mechanism (SMAM) (c) The architecture of nonlinear activation free block (NAFBlock)~\cite{chen2022simple}. (d) Simplified Channel Attention (SCA).
 }

	\label{fig:fir_h}
\end{figure*}

\subsection{Mix Hierarchy Network}
Existing methods for image restoration typically use the following architecture designs: 1). The single-stage encoder-decoder, by using CNN or Transformer~\cite{2018Unprocessing,2018learning,deganv2,Zamir2021Restormer,liang2021swinir} first gradually map the input to low-resolution representations, and then progressively apply reverse mapping to recover the original resolution. Although these models effectively encode multi-scale contextualized information, they tend to sacrifice spatial details.2). Two branches for spatial details and  contextualized information. These networks  divide the task of image restoration into estimating the structure and detail components of the target signals from the input by employing a light-weight sub-network~\cite{Zamir2021MPRNet,Zhang_2019_CVPR,RESCAN}. In this way, spatial details and contextualized information is gradually captured, and the image restoration process becomes more controlled. However, these model parameters are too large, the MACs are high, may yield sub-optimal results, and the loss of $nan$ is easy to occur. Based on the merits of these  structure designs, we propose a mix hierarchy network where earlier use the encoder-decoder network to capture context information and later operates on the full resolution that preserves spatial detail.

\subsection{Encoder-Decoder Subnetwork}
As shown in Figure~\ref{fig:fir_h}(a), we adopt the classic U-shaped architecture with skip-connections as our encoder-decoder subnetwork and add or replace the following components. 
 
\noindent\textbf{Nonlinear Activation Free Block (NAFBlock)}. 
In order to reduce computing resources and speed up computing, we adopt the nonlinear activation free block (NAFBlock)~\cite{chen2022simple} as our base block in our MHNet. Fig~\ref{fig:fir_h}(c) illustrates the process of obtaining an output $Y$ from an input $X$ using Layer Normalization (LN), Convolution, Simple Gate (SG), and Simplified Channel Attention (SCA). Express as follows:
\begin{equation}
\begin{aligned}
	\label{equ:0xnaf}
	X_1 &= X +f_{1 \times 1}^c(SCA(SG(f_{3 \times 3}^{dwc} (f_{1 \times 1}^c(LN(X)))))) 
 \\
 Y &= X_1 +f_{1 \times 1}^c(SG(f_{1 \times 1}^c(LN(X_1))))
    \\
    SCA &= X_{f3} \cdot f_{1 \times 1}^c( GAP(X_{f3}))
    \\
    SG &= X_{f1} \cdot X_{f2} 
\end{aligned}
\end{equation}
where $f_{1 \times 1}^c$ represents $1 \times 1$ convolution, $f_{3 \times 3}^{dwc}$ denotes the $3 \times 3$ depth-wise convolution, and GAP is the global average pooling. For a given input $X_{f0}$, SG initially splits it into two features $X_{f1}, X_{f2} \in \  R^{H \times W \times \frac{C}{2}}$  along channel dimension. Subsequently, SG calculates the $X_{f1}, X_{f2}$ using a linear gate.  For a more intuitive representation, we illustrate $SCA(\cdot)$ in Fig.\ref{fig:fir_h}(d).

\noindent\textbf{Selective Multi-head Attention Mechanism (SMAM)}. Due to the limited receptive field, there are many hindrances by using CNN for image restoration. However, there also are a few challenges when using self-attention in vision, such as the overheads for training. The computation on a global scale results in a quadratic complexity in relation to the number of tokens as shown in Eq.~\ref{equ:10}, rendering it inadequate for the representation of high-resolution images. 
\begin{equation}
	\label{equ:10}
    \mathcal{O}_{MSA} = 4hwC^2 + 2(hw)^2C  
\end{equation}

Furthermore, the conventional self-attention approach necessitates the computation of attention maps for all query-key pairs, posing challenges for image restoration due to potential noisy interactions among irrelevant features.
To this end, we design  a  selective multi-head attention mechanism (SMAM) (see Figure~\ref{fig:fir_h}(b))  as the middle block of the encoder-decoder. SMAM allows for the adaptive retention of the most crucial attention scores, enabling attention to operate on smaller resolutions and thereby reducing training costs. Additionally, taking inspiration from~\cite{Zamir2021Restormer}, we implement SAM across channels rather than the spatial dimension to decrease time complexity.

Specifically, given an input feature $\textbf{F} \in \mathbb R^{H \times W \times C } $, we first apply $1 \times 1$ convolutions and $3 \times 3$  depth-wise convolutions to aggregate channel-wise context, yielding $query, key$ and $value$ matrices $\mathbf{Q} \in \mathbb R^{H \times W \times C}, \mathbf{K} \in \mathbb R^{H \times W \times C}$, $\mathbf{V} \in \mathbb R^{H \times W \times C}$. Then we reshape $\mathbf{Q}, \mathbf{K}, \mathbf{V}$ to $\mathbf{\hat{Q}} \in \mathbb R^{(H W) \times \frac{C}{h} \times h}$, $\mathbf{\hat{K}} \in \mathbb R^{(H W) \times \frac{C}{h} \times h}$, $\mathbf{\hat{V}} \in \mathbb R^{(H W) \times \frac{C}{h} \times h}$, where $h$ is the number of head.
Subsequently, we compute similarities between $\hat{Q}$ and $\hat{K}^T$, selectively choosing the most crucial attention scores for softmax computation. The probabilities of other elements are then replaced with 0. The entire self-attention procedure of the developed SMAM is formulated as:
\begin{equation}
    \label{sam}
    SMAM(\mathbf{\hat{Q}},\mathbf{\hat{K}},\mathbf{\hat{V}})=Softmax(\lambda(\frac{\hat{Q}\hat{K}^T}{\beta})) \hat{V} 
\end{equation}
where $\beta$ is a learning scaling parameter used to adjust the magnitude of the dot product of $\hat{Q}$ and $\hat{K}$ prior to the application of the softmax function. And $\lambda(\cdot)$ is the trainable selection operstor:
\begin{equation}
    \label{sam2}
    \lambda(\gamma) = \begin{cases}
  \gamma & \text{if } \gamma \geq t \\
  0 & \text{oterwise. } 
\end{cases}
\end{equation}

Finally, we reshape the attention matrix back to its original dimensions of $\mathbb R^{H \times W \times C}$.  And the computation change from  quadratic complexity $\mathcal{O}_{MSA}$ to  linear complexity $\mathcal{O}_{SMAM}$.
\begin{equation}
	\label{equ:101}
    \mathcal{O}_{SMAM} = 5hwC^2 + hwC
\end{equation}

\begin{figure*}[!htb] 
	\centering
	\includegraphics[width=1\textwidth]{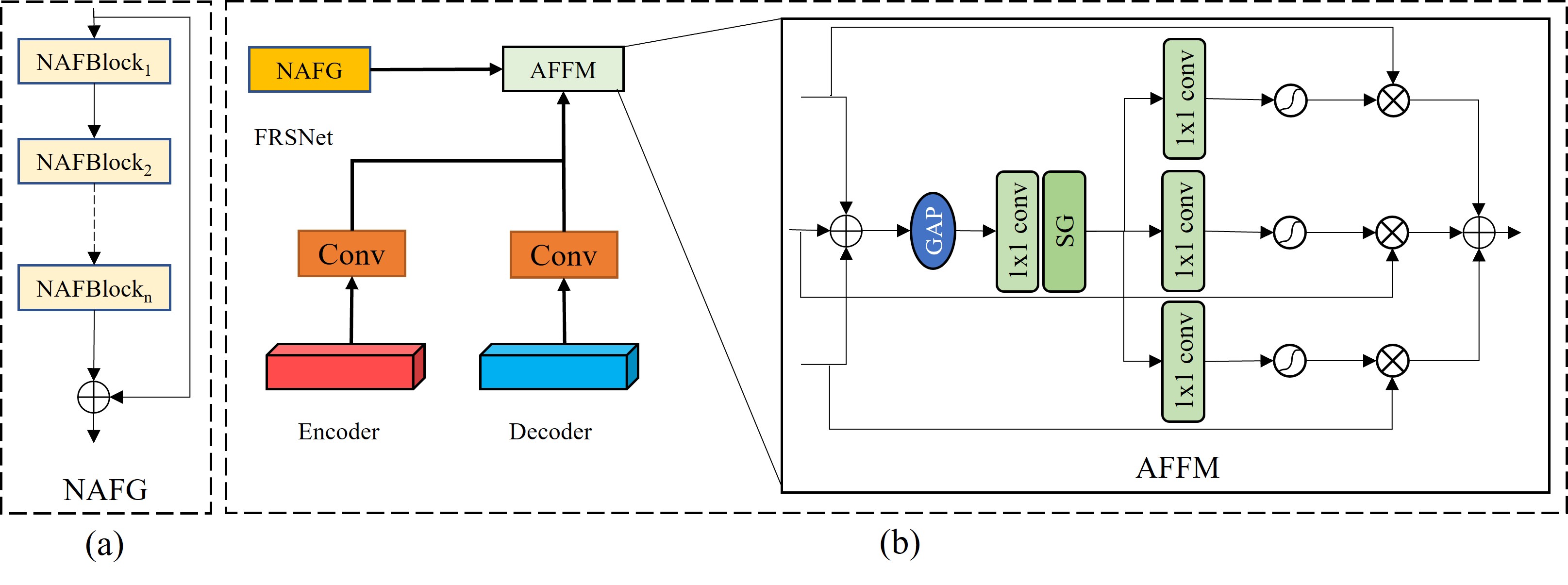}
	\caption{(a) The  architecture of nonlinear activation free block groups (NAFG). Each NAFG further contains multiple nonlinear activation free blocks (NAFBlocks). (b) Adaptive feature fusion mechanism (AFFM) between an encoder-decoder subnetwork and FRSNet.}
	\label{fig:sec_h}
\end{figure*}
\subsection{Full Resolution Subnetwork (FRSNet)}
As we mentioned before, we use the encoder-decoder subnetwork to capture context information. While recovering images demands a complex balance between spatial details and  contextualized information, thus we introduce the full resolution subnetwork (FRSNet)  to preserve fine details from the degraded image to the restored image.  It consists of N numbers nonlinear activation free block groups (NAFGs), each of which further contains NAFBlocks. Figure~\ref{fig:sec_h}(a) shows the architecture of NAFG.  Formally, let given a features  $\mathbf{X_{i}}$ as the input of the $i+1_{th}$ NAFG, $\mathbf{X_{ei}} \in \mathbb R^{\frac{H}{i^2} \times \frac{W}{i^2} \times {i^2}C}$, $\mathbf{X_{di}} \in \mathbb R^{\frac{H}{i^2} \times \frac{W}{i^2} \times {i^2}C}$ be the output in the $i+1$-th level encoder-decoder respectively $(i=1,2,3,4)$. The encoding procedures of NAFG can be defined as:  
\begin{equation}
\begin{aligned}
	\label{equ:NAFG}
   NAFG(X_{i}) &= X_{i} + NAFBlock_{l->1}(X_{i})
    \\
    X_1 &= f_{1 \times 1}^c([X_0, X_{d4}])
    \\
    X_{i+1} &= AFFM(NAFG(X_{i}),  X_{di} , X_{ei})
\end{aligned}
\end{equation}
where $[\cdot]$ represents the channel-wise concatenation, $NAFBlock_{l->1}$ is the l number NAFBlocks, and $AFFM(\cdot)$ denoets the adaptive feature fusion mechanism which will be described below.

\subsection{Adaptive Feature Fusion Mechanism (AFFM)}
The integration of information exchange across various hierarchies stands as a pivotal aspect of our mixed hierarchy architecture. Commonly employed methods~\cite{Zamir2021MPRNet,PREnet,RESCAN} for feature aggregation involve simplistic concatenation or summation. Nevertheless, these options offer limited expressive. Inspired by the~\cite{Zamir2022MIRNetv2,li2019selective}, we propose the adaptive feature fusion mechanism (AFFM) for integrating features from distinct hierarchies. AFFM employs a self-attention mechanism, illustrated in Figure~\ref{fig:sec_h}(b), to selectively combine spatially-precise details and rich contextual information. 
Specifically, given the output features of the encoder-decoder, denoted as $X_{ei}$ and $X_{di}$, along with the output feature of the $i$-th NAFG, denoted as $X_i$, we initially employ PixelShuffle and convolution operations to align the channel count and feature map size of $X_{ei}$ and $X_{di}$ with those of $X_i$. Subsequently, we combine them using an element-wise sum as follows:
\begin{equation}
\begin{aligned}
	\label{equ:s1sum}
 \hat{X_{ei}} &= f_{1 \times 1}^c(PS(X_{ei}))
 \\
 \hat{X_{di}} &= f_{1 \times 1}^c(PS(X_{di}))
 \\
    X_S &= X_i + \hat{X_{ei}} +  f\hat{X_{di}}
\end{aligned}
\end{equation}
where $PS(\cdot$ represents the PixelShuffle to upsamp the $X_{ei}$ and $X_{di}$.

Next, we perform global average pooling (GAP) across the spatial dimensions of $X_S \in \mathbb{R}^{H \times W \times C}$ to compute channel-wise statistics. Subsequently, we apply a channel-downscaling convolution layer and  SG to generate a compact feature representation. This compact feature is then passed through three parallel channel-upscaling convolution layers to create global feature descriptors. This descriptors applies the softmax
function to  yield attention score that we use to adaptively modify the features $X_i$, $X_{ei}$ and $X_{di}$, respectively. Finally, we use element-wise addition to fuse this features to obtain $X_{i+1}$. The entire  procedure of AFFM is  defined as:
\begin{equation}
\begin{aligned}
\label{equ:select}
    X^{c}_S &= SG(f_{1 \times 1}^c(GAP(X_S)))
    \\
    X^{d}_e &= softmax(f_{1 \times 1}^c(X^{c}_S))
    \\
    X^{d}_d &= softmax(f_{1 \times 1}^c(X^{c}_S))
      \\
    X^{d}_x &= softmax(f_{1 \times 1}^c(X^{c}_S))
      \\
    X_{i+1} &= X^{d}_e \cdot \hat{X_{ei}} + X^{d}_d \cdot \hat{X_{di}} +  X^{d}_x\cdot X_i
\end{aligned}
\end{equation}

The proposed AFFM offers several advantages. Firstly, it mitigates feature loss during the up and down sampling  in the encoder-decoder, thereby enhancing feature information at the subsequent hierarchy and promoting more stable model training. Moreover, the dynamic adjustment in the fusion process ensures the integration of the most useful information, enhancing the aggregated features for more effective high-quality image reconstruction.

\begin{table*}
\centering
\caption{Image deraining results. The best and second best scores are \textbf{highlighted} and \underline{underlined}. Our MHNet is better than the state-of-the-art by 0.32 dB. }\label{tb:derain}
\begin{tabular}{ccccccccc||cc}
    \hline
    \multicolumn{1}{c}{} & \multicolumn{2}{c}{Test100~\cite{Test100}}  & \multicolumn{2}{c}{Test1200~\cite{MSPFN}} & \multicolumn{2}{c}{Rain100H~\cite{Rain100}} & \multicolumn{2}{c||}{Rain100L~\cite{Rain100}} & \multicolumn{2}{c}{Average} 
    \\
   Methods &PSNR $\uparrow$ &  SSIM $\uparrow$  & PSNR $\uparrow$ & SSIM $\uparrow$ &PSNR $\uparrow$ &SSIM $\uparrow$ & PSNR $\uparrow$&SSIM $\uparrow$ &PSNR $\uparrow$ & SSIM $\uparrow$
    \\
    \hline\hline
    DerainNet~\cite{DerainNet}  & 22.77 & 0.810  & 23.38  & 0.835  & 14.92 &  0.592  & 27.03 & 0.884 & 22.48 & 0.796 
    \\
     SEMI~\cite{semi}  & 22.35  & 0.788  & 26.05 & 0.822  & 16.56 &  0.486 & 25.03 & 0.842 & 22.88  & 0.744 
     \\
    DIDMDN~\cite{DIDMDN} & 22.56 & 0.818  & 29.65 & 0.901  & 17.35 &  0.524 &25.23 & 0.741 & 24.58  & 0.770
     \\
    UMRL~\cite{UMRL}  & 24.41 &0.829 & 30.55 &  0.910   & 26.01 & 0.832 & 29.18 & 0.923& 28.02  & 0.880 
       \\
    RESCAN~\cite{RESCAN}  & 25.00 & 0.835 & 30.51 & 0.882  &26.36 & 0.786 & 29.80 &0.881 & 28.59  & 0.857
       \\
     PreNet~\cite{PREnet}  & 24.81 &0.851 & 31.36&  0.911   & 26.77 & 0.858  &32.44 & 0.950 &29.42  & 0.897 
    \\
   MSPFN~\cite{MSPFN}  & 27.50 & 0.876 & 32.39 &  0.916   & 28.66 & 0.860  & 32.40 & 0.933 & 30.75  & 0.903
       \\
     MPRNet~\cite{Zamir2021MPRNet}  & 30.27 & 0.907 & 32.91 &  0.916   & 30.51 & 0.890  & 37.20 & 0.965 & 32.73 & 0.921
       \\
     SPAIR~\cite{SPAIR}  & 30.35 & \underline{0.909} & 33.04 &  0.922   & 30.95 & 0.893  & 37.30 & 0.978& 32.91 & \textbf{0.926}
     \\
    
      HINet~\cite{Chen_2021_CVPR}  & 30.29 & 0.906 & 33.05&  0.919   & 30.65 & 0.894 & 37.28 & 0.970 & 32.81  & 0.922
     \\
     U$^2$Former~\cite{u2former} & - & - & \textbf{33.48} & \textbf{0.926} & 30.87 & 0.893 & 39.31 & 0.982 & - &-
     \\
     MDARNet~\cite{MDARNet} & 28.98 & 0.892 & 33.08 & 0.919 & 29.71 & 0.884 & 35.68 & 0.961 & 31.86 &0.914
     \\
      SDLNet~\cite{SDLNet} &- & - &- &- & 30.83 & 0.891 & 39.52 & 0.981 & - &-
     \\
     Restormer~\cite{Zamir2021Restormer} &\textbf{31.32} & \textbf{0.910} & 33.19 & \textbf{0.926} & 31.06 &0.895 &38.99 &0.975 &33.64 & \underline{0.926} 
     \\
     DRSformer~\cite{DRSformer} & - & - & - & - & \textbf{31.13} & \textbf{0.903} &\underline{40.01} & \textbf{0.989} & - & -
     \\
     \hline
      NAFNet~\cite{chen2022simple}  & 30.25 & 0.908 & 32.92 &  0.917   & 30.40 & 0.891  & 37.40 & 0.964 & 32.73  & 0.921
       \\
     \hline
      \textbf{MHNet(Ours)}  & \underline{31.25} & 0.901 & \underline{33.45} &  \underline{0.925}  &\underline{ 31.08} & \underline{0.899}  & \textbf{40.04} & \underline{0.985} & \textbf{33.96} & \textbf{0.928}
    \\
    \hline
\end{tabular}
\end{table*}

\section{Experiments}
We evaluate the proposed MHNet on benchmark datasets  for two image restoration tasks, including \textbf{(a)} image deraining, and \textbf{(b)} image deblurring.

\subsection{Implementation Details}
We train the proposed MHNet without any pre-training and separate models for different image restoration tasks. In the encoder-decoder subnetwork, we employ 4 layers of encoder-decoder and 1 middle block. We apply $[1, 1, 1, 28]$ NAFBlock at each scale of the encoder, $[1, 1, 1, 1]$ NAFBlock at each scale of the decoder, and 8 heads for the SMAM. In the full resolution subnetwork, we use 4 NAFGs, each of which further contains 8 NAFBlocks. Depending on the task complexity, we scale the network width by setting the number of channels to 32 for deraining, and 32, 64 for deblurring. We train models with Adam~\cite{2014Adam} optimizer($\beta_1=0.9, \beta_2=0.999$) and PSNR loss for $4 \times 10^5$ iterations with the initial learning rate $5 \times 10^{-4}$ gradually reduced to $1 \times 10^{-7}$  with the cosine annealing[~\cite{2016SGDR}. We extract patches of size $256 \times 256$ from training images, and the batch size is set to $32$. For data augmentation, we perform horizontal and vertical flips. 


\subsection{Image Deraining Results}
\begin{figure*}[!htb] 
    \centerline{\includegraphics[width=1\textwidth]{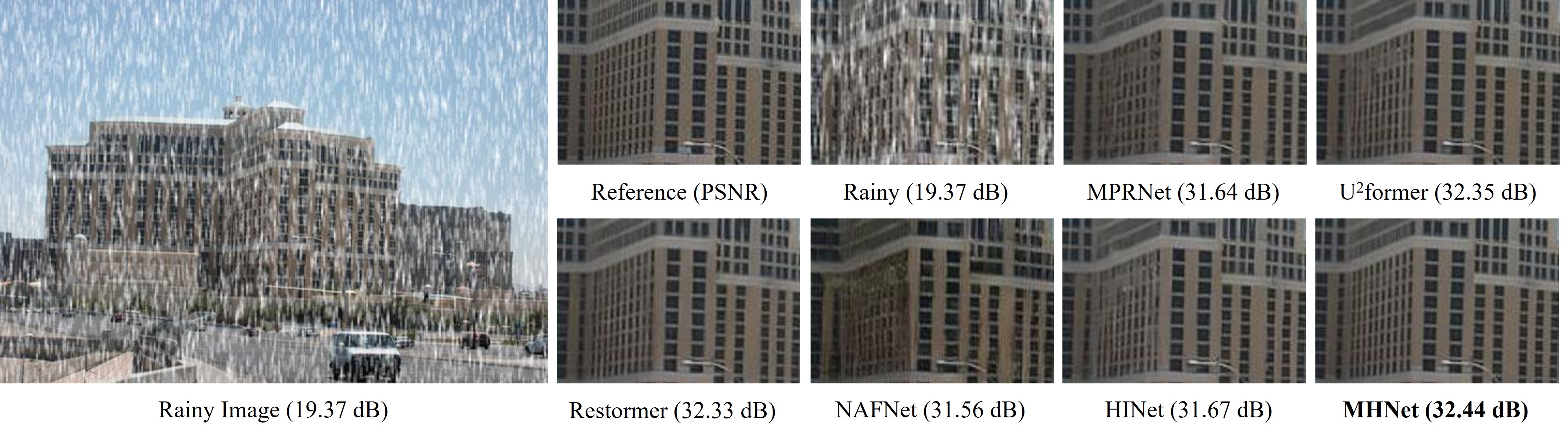}}
	\caption{\textbf{Image deraining} example. The outputs of the MHNet exhibit
no traces of rain streaks on both image sample. MHNet also recovers the most detailed images.}
	\label{fig:05}
\end{figure*}
Following the prior work~\cite{Zamir2021MPRNet,MSPFN,SPAIR}, we computer PSNR/SSIM scores using the Y channel in YCbCr color space for image deraining task. As Table.~\ref{tb:derain} shows, our method gains  consistent and significant performance gains over existing approaches. 
Compared to our baseline network NAFNet~\cite{chen2022simple}, our method achieves 1.23 dB improvement when averaged across all datasets. The improvements on some datasets are as large as 2.64 dB. e.g., Rain100L~\cite{Rain100}. Compared to the recent best algorithm DRSformer~\cite{DRSformer}, we obtain competitive results.  Compared to the Resformer~\cite{Zamir2021Restormer},  our method achieves 0.32 dB improvement when averaged across all datasets.
In comparison to our baseline network, NAFNet~\cite{chen2022simple}, our method demonstrates a notable 1.23 dB improvement on average across all datasets, with some datasets showing improvements as substantial as 2.64 dB (e.g., Rain100L~\cite{Rain100}). Compared to Resformer~\cite{Zamir2021Restormer}, our approach achieves a 0.32 dB improvement on average across all datasets. Furthermore, compared to the state-of-the-art algorithm DRSformer~\cite{DRSformer}, our method yields competitive results.

Figure.~\ref{fig:05} shows our MHNet is effective in removing rain streaks of different orientations and magnitudes while effectively preserving the structural content.

\subsection{Image Deblurring Result}
\begin{figure*} 
    \centerline{\includegraphics[width=1\textwidth]{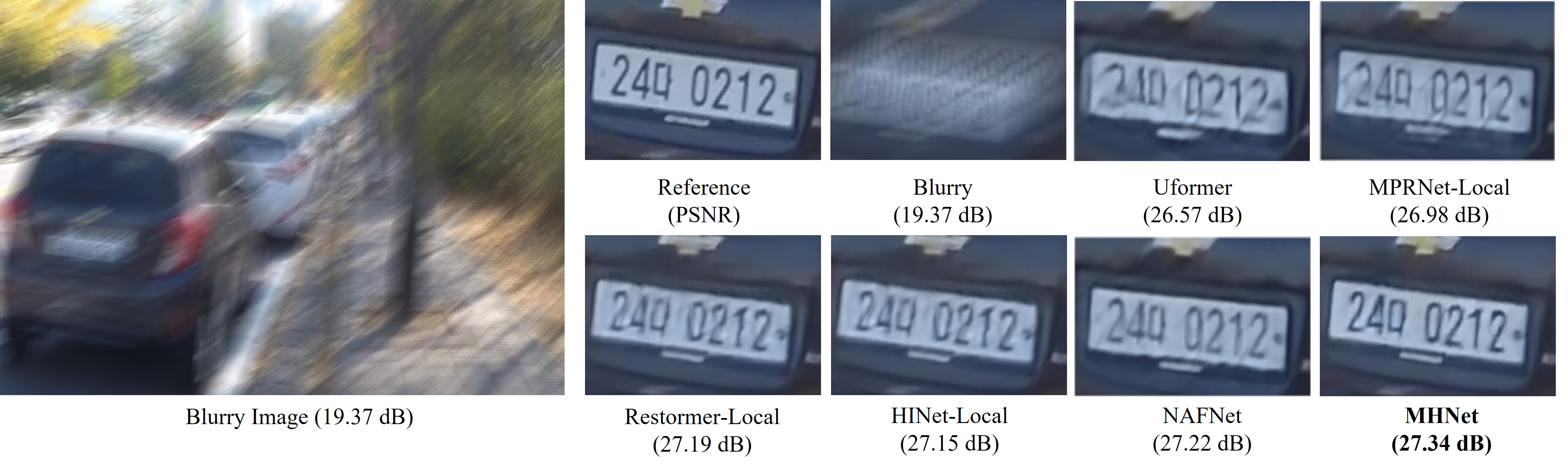}}
	\caption{\textbf{Image deblurring} example on the GoPro dataset~\cite{Gopro}. Compared to the state-of-the-art methods, our MHNet
restores sharper and perceptually-faithful images.}
	\label{fig:06}
\end{figure*}

As  shown in Table.~\ref{tb:deblur}, we evaluated the performance of image deblurring approaches on the GoPro~\cite{Gopro} and HIDE~\cite{HIDE} datasets. Overall, our MHNet outperformed other methods, with a performance gain of 0.32dB when averaging across all datasets~\cite{Gopro,HIDE}. Specifically, compared to our baseline network NAFNet~\cite{chen2022simple}, we improve 0.2 dB and 0.14 dB at 32  and 64  number of channels, respectively.  Compared with previous best  method, Restormer-local~\cite{Zamir2021Restormer}, we improve 0.19 dB on the GoPro~\cite{Gopro} dataset and 0.44 dB on HIDE~\cite{HIDE} dataset. In comparison to the recent algorithm MSFS-Net-Local~\cite{MSFSnet}, our MHNet achieves a 0.47 dB improvement on average across all datasets. Noted that, even though our network is trained solely on the GoPro~\cite{Gopro} dataset, it still achieves state-of-the-art results (31.93 dB in PSNR) on the HIDE~\cite{HIDE} dataset. This demonstrates its impressive generalization capability. Figure.~\ref{fig:06} shows some of the images deblurred by the evaluation method, our model recovered clearer images and was closer to ground truth.

Furthermore, in order to demonstrate the applicability of our approach in road scenarios, we utilized the CamVid and KITTI datasets, both of which are open datasets featuring road scenes, to generate a dataset of motion-blurred image pairs through a random motion blur generation method. The results are shown in Table.~\ref{tb:00200}, our method outperforms the state-of-the-art methods.

\begin{table}[ht]
\centering
\caption{Image deblurring results. The proposed M3SNet is trained only on the GoPro dataset but achieves a 0.32 dB improvement over the state of the art on the average of the effects on both datasets. \label{tb:deblur}}
\resizebox{\linewidth}{!}{
\begin{tabular}{ccccc||cc}
    \hline
    \multicolumn{1}{c}{} & \multicolumn{2}{c}{GoPro~\cite{Gopro}}  & \multicolumn{2}{c||}{HIDE~\cite{HIDE}} & \multicolumn{2}{c}{Average}
    \\
   Methods & PSNR $\uparrow$ & SSIM $\uparrow$ & PSNR $\uparrow$ & SSIM $\uparrow$   &  PSNR $\uparrow$ &  SSIM $\uparrow$
    \\
    \hline\hline
    DeblurGAN~\cite{Degan}  & $28.70 $ & $0.858$ & $24.51$ & $0.871$& 26.61  &0.865 
    \\
    Nah $et al.$~\cite{Gopro} & $29.08$ & $0.914$ & $25.73$ & $0.874$& 27.41   &0.894
    \\
    DeblurGAN-v2~\cite{deganv2} & 29.55 & 0.934 & 26.61 & 0.875& 28.08 &0.905
    \\
    SRN~\cite{2018Scale} & 30.26 & 0.934 & 28.36 & 0.915 &29.31   &0.925
    \\
    Gao $et al.$~\cite{Gao2019DynamicSD} & 30.90 & 0.935 & 29.11 & 0.913 &30.01   &0.924
    \\
    DBGAN~\cite{DBGAN} & 31.10 & 0.942 & 28.94 & 0.915 &30.02 &0.929
    \\
    MT-RNN~\cite{MTRNN} & 31.15 & 0.945 & 29.15 & 0.918 &30.15  &0.932
    \\
    DMPHN~\cite{Zhang_2019_CVPR} & 31.20 & 0.940 & 29.09 & 0.924 & 30.15  &0.932
    \\
    Suin $et al.$~\cite{Suin2020SpatiallyAttentivePN} & 31.85 & 0.948 & 29.98 & 0.930&30.92 &0.939
    \\
    SPAIR~\cite{SPAIR} & 32.06 & 0.953 & 30.29 & 0.931 &31.18  &0.942
    \\
    MIMO-UNet++~\cite{2021Rethinking} & 32.45 & 0.957 & 29.99 & 0.930 &31.22 &0.944
    \\
    MPRNet~\cite{Zamir2021MPRNet} & 32.66 & 0.959 & 30.96 & 0.939 &31.81  &0.949
    \\
    MPRNet-local~\cite{Zamir2021MPRNet} & 33.31 & 0.964 &31.19 &0.945 &32.25 &0.955
    \\
    Restormer~\cite{Zamir2021Restormer} & 32.92 & 0.961 & 31.22 & 0.942 &32.07 &0.952
    \\
    Restormer-local~\cite{Zamir2021Restormer} & 33.57 & 0.966 & \underline{31.49} & 0.945 &\underline{32.53}  &0.956
    \\
    Uformer~\cite{Wang_2022_CVPR} &32.97 & \underline{0.967} &30.83 &\textbf{0.952} &31.90  &\textbf{0.960}
     \\
    HINet~\cite{Chen_2021_CVPR}&32.71&-&-&-&-&-
    \\
    HINet-local~\cite{Chen_2021_CVPR}&33.08&0.962&-&-&-&-
    \\
    MSFS-Net~\cite{MSFSnet} & 32.73 & 0.959 & 31.05 & 0.941& 31.99 & 0.950 
    \\
    MSFS-Net-local~\cite{MSFSnet} & 33.46 & 0.964 & 31.30 & 0.943 & 32.38 & 0.954 
    \\
    \hline
    NAFNet-32~\cite{chen2022simple}&32.83&0.960&-&-&-&-
    \\
    NAFNet-64~\cite{chen2022simple}&\underline{33.62}&\underline{0.967}&-&-&-&-
    \\
    \hline
    \textbf{MHNet-32 (ours)}&33.03&0.965&31.02&0.949&32.03  &\underline{0.957}
    \\
    \textbf{MHNet-64 (ours)}&\textbf{33.76}&\textbf{0.969}&\textbf{31.93}&\underline{0.951}&\textbf{32.85}  & \textbf{0.960}
    \\
    \hline
\end{tabular}
}

\end{table}

\begin{table}[htb]
\begin{center}
\caption{Comparing the image quality achieved by the proposed method with state-of-the-art techniques, we evaluated the performance using metrics such as SNR, PSNR, and SSIM. The term "Blurred" refers to the difference in these metrics between the motion-blurred image and the original ground-truth image.}
\begin{tabular}{cccc}
\\ \hline
Methods                                  & SNR            & PSNR           & SSIM                            \\ \hline
Blurred                                  & 11.86          & 19.87          & 0.57                            \\
DeblurGAN-v2~\cite{deganv2}                             & 13.61          & 21.63          & 0.67                            \\
MPRNet~\cite{Zamir2021MPRNet}                                   & 14.99          & 23.00          & \textbf{0.76}                   \\
HINet~\cite{Chen_2021_CVPR}                                    & 15.15          & 23.16          & \underline{0.75}                      \\
MIMO-UNet++~\cite{2021Rethinking}                              & 14.34          & 22.36          & 0.73                            \\ \hline
\textbf{MHNet(our)} & \textbf{15.42} & \textbf{23.19} & \underline{0.75} \\ \hline
\label{tb:00200}
\end{tabular}
\end{center}

\end{table}

\subsection{Ablation Studies}
The ablation studies are conducted on image deblurring (GoPro~\cite{Gopro}) to analyze the impact of each of our model components. Next, we describe the impact of each component.

\begin{table}
    \centering
    \caption{Ablation study on individual components of the
proposed MHNet.}
\label{tab:abl}
    \begin{tabular}{cccc}
    \hline
         Mix Network& SMAM &AFFM  &PSNR 
         \\
         \hline
         U-Net + U-Net& \faTimes &\faTimes  & 32.66
         \\
         FRSNet + FRSNet& \faTimes &\faTimes  & 31.88
         \\
         U-Net + FRSNet& \faTimes &\faTimes  & 32.74
         \\
         U-Net + FRSNet& \faCheck &\faTimes  & 32.85
         \\
         U-Net + FRSNet& \faTimes &\faCheck  & 32.99
         \\
         U-Net + FRSNet& \faCheck &\faCheck  & 33.03
         \\
         \hline
    \end{tabular}

\end{table}

\textbf{Choices of subnetwork.}
Our goal is to find two subnetwork models to capture spatial details and context information respectively and construct a mixed network to achieve image restoration. Therefore, we tested the effect of using different subnetworks on the experimental results. As the Table~\ref{tab:abl} shown, we using the U-Net + FRSNet leads to better performance (33.03 dB) as compared to employing the same design. This shows that combining the encoder-decoder structure with a high-resolution subnetwork can achieve a balance between spatial details and high-level contextualized information while recovering images. 

\textbf{Adaptive Feature Fusion Mechanism (AFFM).} 
The incorporation of information exchange across diverse hierarchies is a crucial aspect of our mixed hierarchy architecture. We analyze the feature aggregation strategy in Table~\ref{tb:ffm}. It shows that the AFFM achieves improvements of 0.18 dB, 0.16 dB, and 0.12 dB  compared to   summation, concatenation and CSFF~\cite{Zamir2021MPRNet}. We also verify that the effect of feature fusion mechanism for hierarchy architecture. As indicated in Table~\ref{tab:abl}, the removal of the AFFM results in a substantial performance degradation, dropping from 32.99 dB to 32.74 dB.

\begin{table}
\caption{The influence of different  feature fusion mechanism.}
\label{tb:ffm}
    \centering
    \begin{tabular}{ccccc}
         \hline
          & Sum  & Concat  & CSFF~\cite{Zamir2021MPRNet}& AFFM
        \\
        \hline
        PSNR & 32.85 & 32.87 & 32.91& 33.03
         \\
         \hline
    \end{tabular}
\end{table}

\textbf{Selective Multi-head Attention Mechanism (SMAM).} The proposed SMAM as the encoder-decoder middle block  is used to  aggregate the  most crucial information contained in the feature maps captured by convolution. To examine the effect of SMAM, we substitute the  attention mechanism in MHNet with several SOTA approaches, including FASA~\cite{kong2023efficient}, W-MSA~\cite{Wang_2022_CVPR}, and MDTA~\cite{Zamir2021Restormer}. As shown in Table.~\ref{tb:amabl}, compared with FASA, W-MSA, and MDTA, our SMAM achieves improvements of 0.1 dB, 0.16 dB, and 0.04 dB, respectively. When we examine MHNet without SMAM to assess the effect of the proposed SMAM on adaptively retain the most crucial attention scores, we find that our method achieves a 0.11 dB improvement with SMAM as shown in Table~\ref{tab:abl}.
\begin{table}
\caption{The influence of different  attention mechanism.}
\label{tb:amabl}
    \centering
    \begin{tabular}{ccccc}
         \hline
           & FASA~\cite{kong2023efficient}  & W-MSA~\cite{Wang_2022_CVPR}  & MDTA~\cite{Zamir2021Restormer}& AFFM
        \\
        \hline
        PSNR  &32.93 &32.87 &32.99  & 33.03
         \\
         \hline
    \end{tabular}
\end{table}

\subsection{Resource Efficient}

\begin{table}
\centering
\caption{The evaluation of model computational complexity. This is conducted with an input size of $256 \times 256$, on an NVIDIA 1060 GPU. \label{tb:04}}

\begin{tabular}{ccccc}
\hline
\multicolumn{2}{c}{Method} & \multicolumn{1}{c}{PSNR} & \multicolumn{1}{c}{Params(M)} & \multicolumn{1}{c}{MACs(G)} 
\\
\hline\hline
\multicolumn{2}{c}{MIMO-UNet++~\cite{2021Rethinking}}  & \multicolumn{1}{c}{$32.68$} & \multicolumn{1}{c}{$16.1$}  & \multicolumn{1}{c}{1235} 
\\
\multicolumn{2}{c}{MPRNet~\cite{Zamir2021MPRNet}}  & \multicolumn{1}{c}{$32.66$} & \multicolumn{1}{c}{20.1}  & \multicolumn{1}{c}{778.2}  
\\
\multicolumn{2}{c}{MPRNet-local~\cite{Zamir2021MPRNet}}  & \multicolumn{1}{c}{$33.31$} & \multicolumn{1}{c}{20.1}  & \multicolumn{1}{c}{778.2}  
\\
\multicolumn{2}{c}{HINet~\cite{Chen_2021_CVPR}}  & \multicolumn{1}{c}{$32.77$} & \multicolumn{1}{c}{88.7}  & \multicolumn{1}{c}{170.7} 
\\
\multicolumn{2}{c}{Restormer~\cite{Zamir2021Restormer}}  & \multicolumn{1}{c}{$32.92$} & \multicolumn{1}{c}{26.13}  & \multicolumn{1}{c}{140} 
\\
\multicolumn{2}{c}{Restormer-local~\cite{Zamir2021Restormer}}  & \multicolumn{1}{c}{$33.57$} & \multicolumn{1}{c}{26.13}  & \multicolumn{1}{c}{140} 
\\
\multicolumn{2}{c}{Uformer~\cite{Wang_2022_CVPR}}  & \multicolumn{1}{c}{$32.97$} & \multicolumn{1}{c}{50.88}  & \multicolumn{1}{c}{89.5} 
\\
\hline
\multicolumn{2}{c}{NAFNet-32~\cite{chen2022simple}}  & \multicolumn{1}{c}{$32.83$} & \multicolumn{1}{c}{17.1}  & \multicolumn{1}{c}{32}
\\
\multicolumn{2}{c}{NAFNet-64~\cite{chen2022simple}}  & \multicolumn{1}{c}{$33.62$} & \multicolumn{1}{c}{68}  & \multicolumn{1}{c}{65}
\\
\hline
\multicolumn{2}{c}{\textbf{MHNet-32 (ours)}}  & \multicolumn{1}{c}{$33.03$} & \multicolumn{1}{c}{16.9}  & \multicolumn{1}{c}{32}
\\
\multicolumn{2}{c}{\textbf{MHNet-64 (ours)}}  & \multicolumn{1}{c}{$33.76$} & \multicolumn{1}{c}{67}  & \multicolumn{1}{c}{67}
\\
\hline
\end{tabular}
\end{table}
Many deep learning models are now becoming deeper and more complex to achieve higher accuracy. Although large models work better than small models, they are also very resource-intensive. Therefore, it is very important to design a lightweight image restoration model with high accuracy. In our work, we design a mixed hierarchy network that obtains higher accuracy with lower computing resources. 

As the Table.~\ref{tb:04} shown, our MHNet gains the best performance. MIMO-UNet++~\cite{2021Rethinking} has a smaller number of parameters, while MHNet using significantly fewer computational resources with MACs  approximately 40 times smaller than that of MIMO-unet++.  Figure~\ref{fig:param} clearly demonstrates that our MHNet outperforms DRSformer~\cite{DRSformer} by achieving a state-of-the-art result while remarkably reducing the cost reduction by 85\%. Furthermore, when compared to our baseline NAFNet~\cite{chen2022simple}, we achieve superior PSNR values using an equivalent  computational resources. These findings highlight the highly efficient of MHNet. Moreover, through the scaling up of the model size, our MHNet achieves even better performance, highlighting the scalability of MHNet.

\section{Conclusion}
Image restoration usually demands a complex balance between spatial details and high-level contextualized information while recovering images. In this paper, we propose a mixed hierarchy architecture that balances the competing goal by mixing two subnetworks concerned with contextual information and spatial details. Specifically, we design a selective multi-head attention mechanism as the middle block of the encoder-decoder subnetwork to aggregate the most crucial information
contained in the feature maps captured by convolution. To ensure the information exchange between different hierarchies, we propose a adaptive feature fusion mechanism  to selectively combine spatially-precise details and rich contextual information. In addition, to keep the size of our model lightweight and have computational efficiency, we replace or remove the nonlinear activation function with multiplication. Extensive experiments on numerous benchmark datasets demonstrate that MHNet achieves significant performance improvements with low computing resources.

\bibliographystyle{IEEEtran}
\bibliography{egbib}

\vfill

\end{document}